\documentclass[conference]{IEEEtran}
\IEEEoverridecommandlockouts

\usepackage{cite}
\usepackage{amsmath,amssymb,amsfonts}
\usepackage{algorithmic}
\usepackage{graphicx}
\usepackage{textcomp}
\usepackage[dvipsnames]{xcolor}
\def\BibTeX{{\rm B\kern-.05em{\sc i\kern-.025em b}\kern-.08em
    T\kern-.1667em\lower.7ex\hbox{E}\kern-.125emX}}
\usepackage{graphicx}
\usepackage{subcaption}
\graphicspath{ {images/} }
\usepackage[utf8]{inputenc}
\usepackage{fancyhdr}
\usepackage{dirtytalk}
\usepackage{tabularx}
\usepackage[]{nohyperref}
\usepackage{url}
\usepackage[normalem]{ulem} %to strike the words
\usepackage[ruled,vlined]{algorithm2e}

\usepackage{enumitem,kantlipsum}
\usepackage{ulem}

\usepackage{svg}

\usepackage{caption}

\usepackage{xcolor}
\usepackage{svg}
\usepackage{graphicx}
\graphicspath{ {images/} }
\usepackage[utf8]{inputenc}
\usepackage{fancyhdr}
\usepackage{dirtytalk}
\usepackage{tikz}
\usetikzlibrary{shapes.geometric, arrows}
\usepackage{pgfplots}
\pgfplotsset{width=8cm,compat=1.16}
\usepgfplotslibrary{external}
\usepackage[nolist]{acronym}
\begin{acronym}
\acro{LM}{language model}
\acro{LLM}{large language model}
\acro{SLM}{small language model}
\acro{RAG}{retrieval-augmented generation}
\acro{LM}{language model}
\acro{LoRA}{Low-Rank Adaptation}
\acro{QLoRA}{Quantized Low-Rank Adaptation}
\acro{SBERT}{Sentence BERT}
\acro{ML}{machine learning}
\acro{TF-IDF}{term frequency and inverse document frequency}
\acro{OOD}{out-of-distribution}
\acro{QnA}{questions and answering}
\acro{AI}{artificial intelligence}

\acro{TeleQuAD}{telecom question answering dataset}
\acro{3GPP}{Third Generation Partnership Project}
\acro{O-RAN}{open radio access networks}
\acro{MCQ}{multiple choice question}
\acro{RoPE}{rotary position embedding}
\end{acronym}

\tikzstyle{startstop} = [rectangle, rounded corners, minimum width=2cm, minimum height=0.5cm,text centered, draw=black, fill=red!30]
\tikzstyle{process} = [rectangle, minimum width=2cm, minimum height=0.5cm, text centered, draw=black, fill=orange!30, align=left]
\tikzstyle{decision} = [diamond, minimum width=1.0cm, minimum height=0.4cm, text centered, draw=black, fill=green!30]
\tikzstyle{arrow} = [thick,->,>=stealth]

\newcolumntype{L}{>{$}l<{$}}

\usepackage[labelformat=simple]{subcaption}

\begin{document}
\captionsetup[figure]{labelfont={bf},labelformat={default},labelsep=period,name={Figure}}

% \title{The Power Of Small Language Models With RAG In Telecom Systems}
% \title{An Oracle for 3GPP Standards: \\Phi2-based RAG System with LoRA Fine-Tuning and Enhanced Long-Context Support}
\title{Leveraging Fine-Tuned Retrieval-Augmented Generation with Long-Context Support: \\For 3GPP Standards}

% \title{An Oracle for Telecom Networks:\\Leveraging Fine-Tuned Retrieval-Augmented Generation with Long-Context Support}

\author{
    \IEEEauthorblockN{Omar~Erak$^{1*}$, Nouf~Alabbasi$^{1*}$, Omar~Alhussein$^{1}$, Ismail~Lotfi$^{1}$, Amr~Hussein$^{1}$, \\ Sami Muhaidat$^{2,3}$, Merouane~Debbah$^{2}$ }
    \IEEEauthorblockA{$^{1}$KU 6G Research Centre, Department of Computer Science, Khalifa University, Abu Dhabi, UAE\\
                      $^{2}$KU 6G Research Centre, Department of Computer and Information Engineering, Khalifa University, Abu Dhabi, UAE\\
                      $^{3}$Department of Systems and Computer Engineering, Carleton University, Ottawa, Canada\\
                      Emails: omarerak@ieee.org, Nouf.Alabbasi@ieee.org, omar.alhussein@ku.ac.ae, ismail.lotfi@ku.ac.ae, \\100059484@ku.ac.ae, muhaidat@ieee.org, merouane.debbah@ku.ac.ae}
}

% \author{
% Nouf~Alabbasi,~\IEEEmembership{Student~Member,~IEEE,}
%         Omar~Erak,~\IEEEmembership{Student~Member,~IEEE,}
%         Ismail~Lotfi,~\IEEEmembership{Member,~IEEE,} \\ Amr~Hussein,~\IEEEmembership{Student~Member,~IEEE,}\\
% Omar~Alhussein,~\IEEEmembership{Member,~IEEE,}\\ 
% and Sami~Muhaidat,~\IEEEmembership{Senior~Member,~IEEE,}
%         and Merouane~Debbah,~\IEEEmembership{Fellow,~IEEE}
% 				\thanks{O. Alhussein is with the KU 6G Research Center, Department of Computer Science, Khalifa University, Abu Dhabi, UAE (e-mail: omar.alhussein@ku.ac.ae).}
%                     \thanks{S. Muhaidat is with the KU 6G Research Center, Department of Computer and Information Engineering, Khalifa University, Abu Dhabi, UAE, and with the Department of Systems and Computer Engineering, Carleton University, Ottawa, ON K1S 5B6, Canada, (e-mail: muhaidat@ieee.org).}
% 				\thanks{Ning~Zhang is with the Department of Electrical and Computer Engineering, University of Windsor, Canada (e-mail: {ning.zhang}@uwindsor.ca).}
%                     \thanks{Weihua~Zhuang is with the Department of Electrical and Computer Engineering, University of Waterloo, (e-mail: {wzhuang}@uwaterloo.ca).}
		        
% 		\vspace{-0.7cm}		}

% Omar Erak, Nouf Alabbasi, Omar Alhussein, Ismail Lotfi, Amr Hussein, Sami Muhaidat, Merouane Debbah
  
\maketitle
\def\thefootnote{*}\footnotetext{These authors contributed equally to this work.}

\begin{abstract}
Recent studies show that large language models (LLMs) struggle with technical standards in telecommunications. We propose a fine-tuned retrieval-augmented generation (RAG) system based on the Phi-2 small language model (SLM) to serve as an oracle for communication networks.
Our developed system leverages forward-looking semantic chunking to adaptively determine parsing breakpoints based on embedding similarity, enabling effective processing of diverse document formats. To handle the challenge of multiple similar contexts in technical standards, we employ a re-ranking algorithm to prioritize the most relevant retrieved chunks.
Recognizing the limitations of Phi-2's small context window, we implemented a recent technique, namely SelfExtend, to expand the context window during inference, which not only boosts performance but also accommodates a wider range of user queries and design requirements from customers to specialized technicians.
For fine-tuning, we utilize the low-rank adaptation (LoRA) technique to enhance computational efficiency during training and enable effective fine-tuning on small datasets. Our comprehensive experiments demonstrate substantial improvements over existing question-answering approaches in the telecom domain, achieving performance that exceeds larger language models such as GPT-4 (which is about 880 times larger in size).
This work presents a novel approach to leveraging SLMs for communication networks, offering a balance of efficiency and performance. This work can serve as a foundation towards agentic language models for networks.
\end{abstract}

\acresetall


\section{Introduction}

LLMs have demonstrated impressive capabilities, from basic automation to complex decision-making \cite{bubeck2023sparks}. They have proven their efficiency in a variety of tasks, including \ac{QnA}, code generation, and other problems.
%and under-graduate level control problems \cite{kevian2024capabilitiesLLMs}. 
Their ability to process natural language and generate human-like responses makes them powerful tools in many fields, including telecommunications. 

As telecom networks grow more complex and data-driven, \acp{LLM} offer significant potential to enhance automation, optimize network management, and improve customer experiences \cite{maatouk2024large, zhang2024interactive, largelanguagemodeldrivencurriculum}. However, to fully leverage agentic LLM-based models in telecommunications, we need to develop models that deeply understand the nuances of telecom systems and possess comprehensive knowledge of telecom models. Building such specialized models is crucial for adapting \acp{LLM} to agentic roles where they can autonomously handle involved tasks, such as dynamic network optimization and predictive maintenance.

LLMs can be adapted to various tasks through fine-tuning or \ac{RAG}.
Fine-tuning enhances the performance of \acp{LLM} by adjusting the model's internal knowledge through iterative training on specialized datasets. However, the telecom industry is rapidly evolving, rendering fine-tuning an expensive and inefficient approach that cannot easily keep up with such a fast-paced industry. Furthermore, once a model is fine-tuned, editing or forgetting a specific piece of information becomes challenging \cite{yao2024largelanguagemodelunlearning}.

\Ac{RAG}, on the other hand, augments text generation with information retrieval, enabling models to produce more accurate and contextually aware responses. This approach allows flexible model adaptation and rapid integration of new information. \ac{RAG} also grounds the model's response in the relevant retrieved context, reducing (yet not entirely eliminating) the risk of hallucination~\cite{huang2023_survey_hallucination_LLMs}. Telecom applications can benefit from real-time data to produce more accurate and up-to-date responses. 
Therefore, we believe LLM-based RAG systems can better enable emerging applications, such as dynamic network management, customer support, and predictive maintenance.

% SML -> a good fit for edge use-cases
Integrating \ac{RAG} into telecommunication systems involves deploying \ac{LLM} frameworks on user equipment and edge devices, a process which presents a significant challenge due to the involved computational intensity of \acp{LLM}, both in terms of training and inference costs. 
This challenge highlights the appeal of SLMs. These \acp{LM} offer computational and storage efficiency while maintaining adequate performance, suggesting their suitability for deployment on edge devices and possible enablement of on-device artificial intelligence (AI) \cite{piovesan2024_Telecom_LLMs_large}.
Several \acp{SLM} have been proposed in the literature, including Microsoft's Phi-2 with 2.7B parameters\cite{PHI2} and \texttt{Gemini Nano 2} with 3.2B parameters~\cite{gemini}. The Phi-2 model is currently considered as the state-of-the-art \ac{SLM} as it is able to match or outperform models up to 25x larger such as \texttt{Llama-2-70b}  which has 70B parameters \cite{llama}.

Recent studies indicate that while state-of-the-art \acp{LLM} perform well on general telecommunications queries, they struggle with questions related to technical standards in the field~\cite{maatouk2023teleqna, bornea2024_telco_RAG}. We believe this is mainly due to the fact that standard-type knowledge and system specifications do not exist in common research papers and other publications, which serve as the main learning sources for \acp{LLM}.
%For instance, GPT-4 has been shown to have limited performance in the \ac{3GPP} standard specifications task in the TeleQnA dataset~\cite{maatouk2023teleqna}.
%A major issue with \ac{QnA} tasks is their limited performance when encountering domain-specific keywords and abbreviations.
For instance, with respect to telecom systems, the polysemy of abbreviations (e.g., SAP: service access point vs. session announcement protocol) can hinder the model from inferring a correct answer. Additionally, the \acp{LM} training on generalized knowledge can interfere with its performance in specialized domain tasks. For instance, certain protocols and methods in telecom-specific domains do not necessarily follow the generally followed-upon practices in broader contexts. A model trained on generalized knowledge may not adequately capture these domain-specific nuances and practices.
%This is because the model has been primarily exposed to common language usage, which can lead to misunderstandings or misinterpretations of terms that have different meanings in specific fields. 

We propose a carefully developed Phi-2-based fine-tuned RAG system to serve as an oracle for communication networks. To the best of our knowledge, this is the first work to present a fine-tuned RAG system for communication networks. Previous works focus on presenting a frozen RAG framework or fine-tuning an \ac{LM}. Our RAG system leverages a forward-looking semantic chunking (or parsing) strategy that adaptively determines breakpoints between sentences based on embedding similarity. This approach enables the system to effectively process documents with diverse formatting. In the \ac{3GPP} documents, a query can often relate to multiple similar contexts, as discussions and paragraphs on related topics may appear in various sections or be phrased similarly. Therefore, we utilize a re-ranking algorithm to further rank the retrieved chunks based on their relevance to the input query.

Since Phi-2 is an \ac{SLM}, its performance is limited by its small context window, rendering it inefficient for certain tasks such as responding to open-ended or under-specified queries. This limitation is particularly relevant in telecom applications, where users can range from customers to specialized technicians. Therefore, we utilize a new technique, namely SelfExtend \cite{self_extend}, to significantly extend the context window during inference. 

Finally, we use \ac{LoRA} not only to enhance computational efficiency during training but also because it allows users to effectively fine-tune on small datasets. Our in-depth experiments demonstrate considerable improvements over existing QnA approaches in telecom, contributing to the ongoing advancement of the field.

The remainder of the paper is organized as follows. Section~\ref{section_related_works} provides an overview of related works. Section~\ref{section_system_model} discusses the proposed fine-tuned Phi-2 RAG system. Section~\ref{section_simulations} provides our experimental results, and Section~\ref{section_conclusions} concludes the paper and discusses some insightful future research directions.

\section{Related Works}\label{section_related_works}

Several benchmarking datasets have been proposed to enable \ac{LLM}-based systems in the telecommunication domain, with mainly three different tasks: text classification, text summarization, and \acp{MCQ}.
In~\cite{karim2023_SPEC5G}, the SPEC5G dataset was introduced with the objective of performing text classification and text summarization in telecom domain. Ericsson's team introduced TeleQuAD, a private 4,000 entry \ac{QnA} dataset, and developed a proprietary TeleRoBERTa, a 124M bidirectional encoder representation from transformers (BERT)-based RAG system\cite{karapantelakis2024_TeleRoBERTa}.

In \cite{maatouk2023teleqna}, the introduction of the TeleQnA dataset marks a significant advancement in evaluating \ac{QnA} tasks for telecommunications.
The TeleQnA dataset contains 10,000 \acp{MCQ} about telecommunication systems, curated from various sources such as 3GPP and research papers. The dataset is verified by telecom human-in-the-loop experts.
% \ac{3GPP} documentation 
Another dataset, namely TSpec-LLM, is recently released~\cite{nikbakht2024_TSPEC}. The authors develop an automated framework to generate QnA pairs from 3GPP specifications, then test a naive RAG architecture to assess the quality of their dataset.
In~\cite{gajjar2024_ORAN_Bench_13K}, Gajjar et al. introduce ORAN-Bench-13K, a dataset dedicated to the evaluation of \ac{O-RAN} tasks. The dataset is based on 116 \ac{O-RAN} specification documents and contains 13,000 pairs of \acp{MCQ}, based on which ORANSight is developed, a \ac{RAG}-based framework.

The work in~\cite{ahmed2024linguisticintelligencelargelanguage} evaluates the performance of various zero shot \acp{LLM} in a few tasks in the telecommunications domain, including a \ac{QnA} task. Among their findings, they note that though \ac{LLM} models such as Zephyr, and Mistral perform outstandingly in the tasks, their performance still comes strikingly short when compared to GPT-3.5 or GPT-4.
% 
% In~\cite{ahmed2024linguisticintelligencelargelanguage}, Ahmed et al. compare the performance of several \acp{SLM} on different tasks specific to the telecommunication domain, including \ac{QnA}, against that of an \ac{LLM} fine-tuned on telecommunication datasets.
% Their major finding is that off-the-shelf \acp{SLM} struggle to achieve the performance of \acp{LLM} specialized on telecommunication datasets.
% 
The authors in~\cite{bariah2023understandingtelecomlanguagelarge} demonstrate the effectiveness of fine-tuning different \acp{SLM} for the telecom domain. 
% The work demonstrates the importance of fine-tuning \acp{SLM} for specialized domains.
% However, although their work outperforms off-the-shelf \acp{SLM}, the absence of RAG integration in their system may limit the scalability and overall performance of their approach.
% 
% 
% 
% They note that smaller models (7b) outstanding
% "zero-shot evaluation of various LLMs within the telecommunications domain, revealing their capabilities and limitations across multiple tasks."
% 
% demonstrate the effectiveness of fine-tuning different \acp{LLM} for the telecom domain. The work demonstrates the importance of fine-tuning \acp{SLM} for specialized domains.
% "While supervised fine-tuned language models or zero-shot LLMs have shown impressive results in one or more tasks within the telecommunications field, a thorough evaluation of the capabilities and limitations of LLMs across this domain remains unexplored."
% LLM 
% 
% However, although their work outperforms off-the-shelf \acp{SLM}, the absence of RAG integration in their system may limit the scalability and overall performance of their approach.
% 
The work in~\cite{bornea2024_telco_RAG} proposes Telco-RAG, a framework specialized for \ac{MCQ} answering for telecom applications, tailored to the specific requirements of telecom standards, particularly \ac{3GPP} documents. Their contribution focuses on modifying the \ac{RAG} framework by employing a router,
using generated candidate answers to enhance retrieval quality, and appending the definitions of acronyms and technical terms to the user's query.
% "using candidate answers in the retrieval process."

% fusion retriever.

To the best of the authors' knowledge, this is the first work to present a \ac{RAG} architecture with fine-tuned SLM generator. Additionally, critical components such as SelfExtend for handling long contexts, re-ranking for enhancing retrieval accuracy, and semantic chunking to preserve contextual coherence have not been proposed in the relevant literature.

\section{Proposed architecture}\label{section_system_model}
\begin{figure*}[htbp]
    \centering
    \includegraphics
    [width=\linewidth]{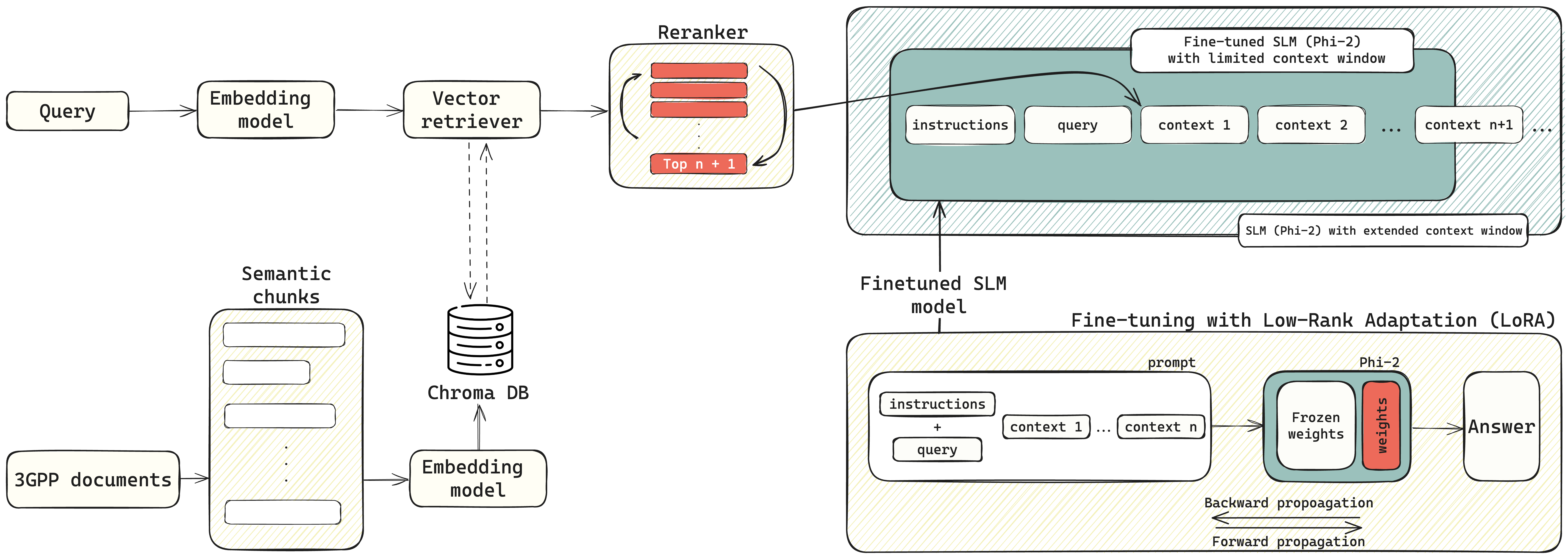}
    \caption{Overview of proposed RAG architecture with semantic chunking, extended context support, and fine-tuned Phi-2 SLM integration for \ac{3GPP} document processing.}
    \label{fig:arch}
\end{figure*}

% \begin{figure*}[htbp]
%     \centering
%     \includegraphics
%     [width=\linewidth]{converted/arch_v4.pdf}
%     \caption{Overview of proposed RAG architecture with semantic chunking, extended context support, and fine-tuned Phi-2 SLM integration for \ac{3GPP} document processing.}
%     \label{fig:arch}
% \end{figure*}

% \begin{figure*}[htbp]
%     \centering
%     % \includegraphics[width=\linewidth]{figures/example_hybrid_retriever_v1.png}
%     \includesvg[scale=0.2]{figures/arch_v2.1.svg}
%     % \includesvg[inkscapelatex=false, scale=0.6, keepaspectratio, svgpath = figures/]{example_hybrid_retriever_v2.svg}

%     \caption{temp example why hybrid retriever is helpful}
%     \label{fig:hybrid_retreiver}
% \end{figure*}

\subsection{General Overview}
\ac{RAG} integrates four key components: a chunking mechanism to segment information, an embedding model to encode the information in a latent space, a retriever to fetch relevant context, and a generator to produce responses.
% \Ac{RAG} is an efficient approach that operates by retrieving a few context chunks for a query, which are then passed, along with the query, to the generator for response generation. The embedding model and database, the retriever, and the generator are the main components in a \ac{RAG} framework.
% There are a few basic components that make up the framework: the embedding model and database, the retriever, and the generator.  
In this work, we present an advanced \ac{RAG} framework, illustrated in Fig. \ref{fig:arch}, that optimizes each component to enhance performance and adapt the model to the telecom domain.
% each component in the pipeline to adapt the model to the telecom domain and enhance performance.
% Our proposed \ac{RAG} framework is illustrated in Fig~\ref{fig:arch}. 
In our architecture, 3GPP documents are first chunked using a semantic chunker then embedded and stored in a vector database (DB). During inference, the user's query is embedded and then passed to the vector retriever.
% hybrid retriever, which consists of a vector and keyword-based retriever. 
The retriever performs a vector similarity search in embedding space to return the nearest relevant neighbors from the indexed corpus.
The retrieved chunks are then passed to a re-ranking algorithm that returns an ordered sublist containing the most relevant chunks.

The Phi-2 generator model is then given the chunks, the query, and a set of instructions to provide an answer. Here, we efficiently fine-tune the generator using \ac{LoRA}. The model initially is fine-tuned with $n$ contexts based on retrieved information and the prompt. During inference, we employ the self-extend technique to dynamically expand the context window, accommodating retrieved contexts beyond the initial $n$ limit.
In what follows, we highlight the key components of the proposed framework.

\subsection{Semantic Chunking Strategy}
% need to add in results

\begin{figure}[htbp]
    \centering
    \includegraphics
    [ width=\linewidth]{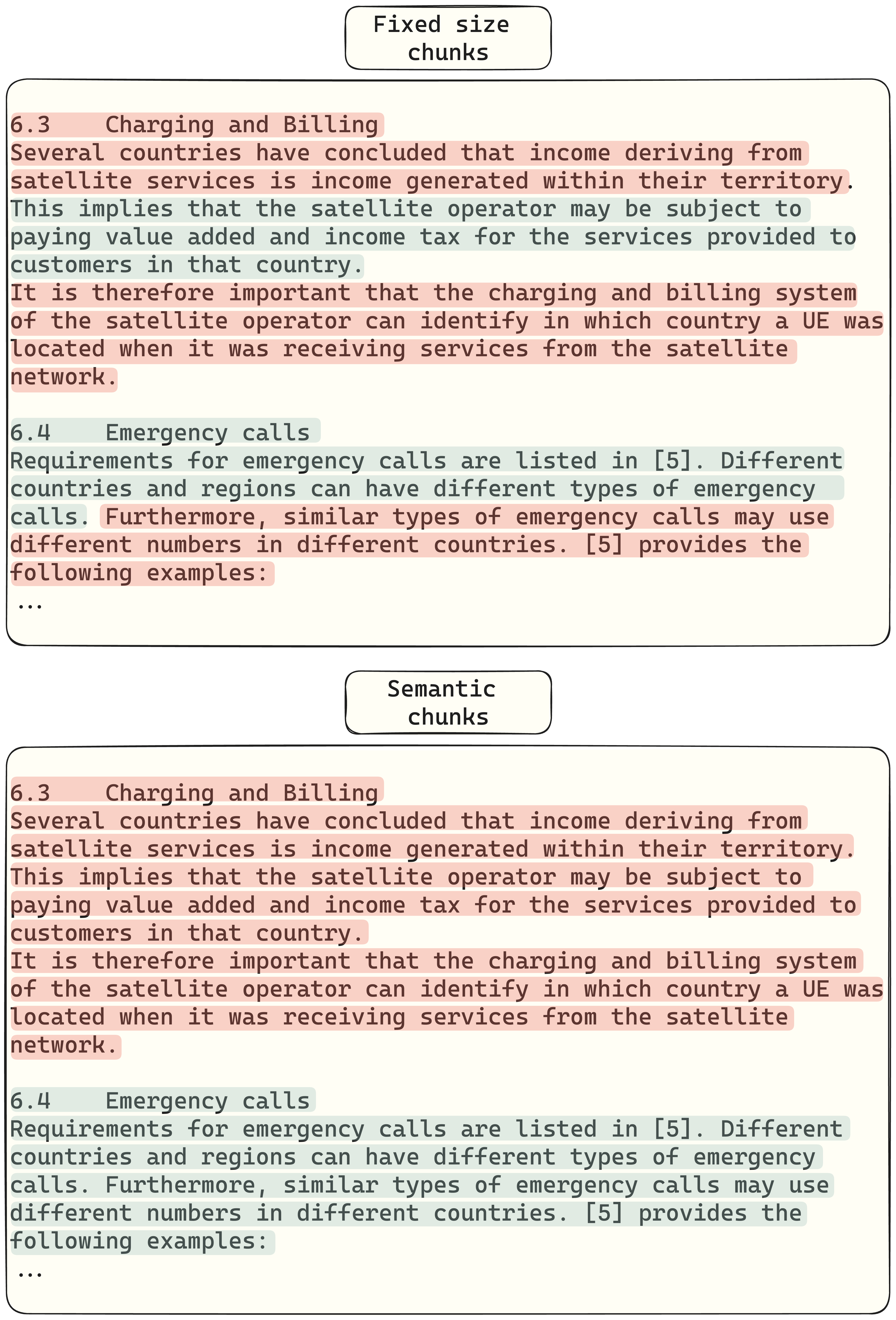}
    \caption{Comparison of fixed-size chunking and semantic chunking applied to an excerpt from a 3GPP document.}
    \label{fig:sem_chunking}
\end{figure}

% \begin{figure}[htbp]
%     \centering
%     % width=\linewidth 
%     \includegraphics
%     [ width=\linewidth]{converted/sem6.pdf}
%     \caption{Comparison of fixed-size chunking and semantic chunking applied to an excerpt from a 3GPP document.}
%     \label{fig:sem_chunking}
% \end{figure}

% % need to add in results
\begin{figure}[htbp]
    \centering
    \includegraphics
    [width=\linewidth]{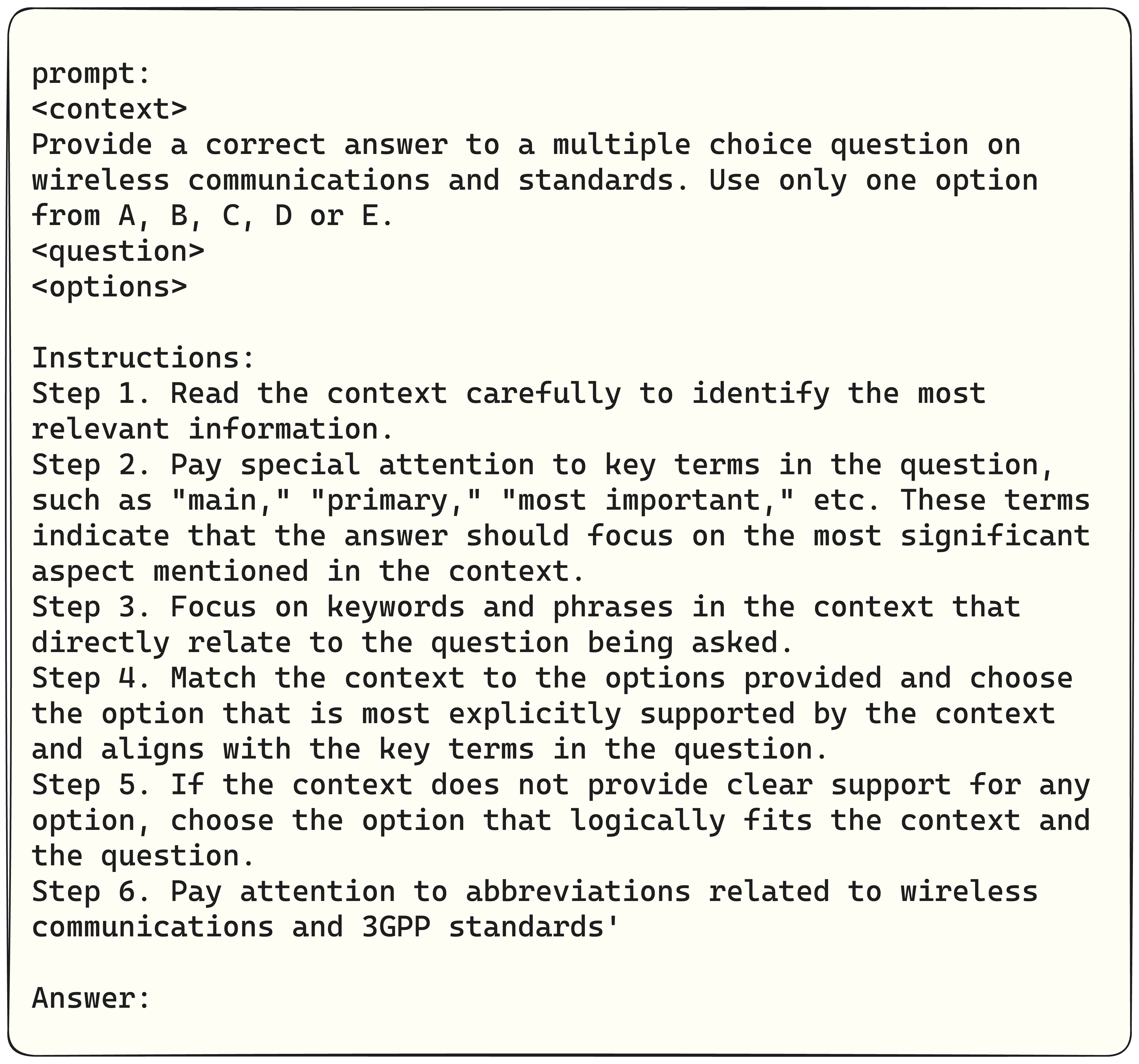}
    \caption{Prompt structure that includes retrieved context and instructions.}
    \label{fig:prompt}
\end{figure}

% \begin{figure}[htbp]
%     \centering
%     \includegraphics
%     [width=\linewidth]{converted/prompt_v4 copy.pdf}
%     \caption{Prompt structure that includes retrieved context and instructions.}
%     \label{fig:prompt}
% \end{figure}

The method of splitting or chunking is critical as improper chunks can lead to inaccurate representations in the embedding space. 
Chunking can be implemented through either fixed-size segments or adaptable segments based on specific criteria. While fixed-size chunking can yield reasonable results and is computationally efficient, it often creates blocks of text that do not consider the content or context. To mitigate this issue, we leverage semantic chunking to split the 3GPP documents \cite{Liu_LlamaIndex_2022}.
% \footnote{https://docs.llamaindex.ai/en/stable/examples/node\_parsers/\\semantic\_chunking/}. 
Semantic chunking adaptively determines breakpoints between sentences based on embedding similarity. 
This way the meaning of the text is preserved through the logical breaks where sentences are semantically connected, rather than arbitrarily cutting the text at fixed intervals. Fig. \ref{fig:sem_chunking} shows an illustrative example from a 3GPP document. 

% This method enhances the 
% preservation of meaning by creating logical breaks 
Given that telecom documents don't follow a strict format, leveraging semantic chunking preserves essential context while minimizing information fragmentation and irrelevant grouping.

\subsection{Embedding Model}
We utilize \textit{bge-small-en-v1.5}, an open-source  embedding model~\cite{bge_embedding}. It is optimized for balancing efficiency and accuracy in text embedding tasks. The model is trained using contrastive learning on a large-scale dataset~\cite{chen2020_contrastive} and it creates vector representations that capture semantic relationships between elements. This method effectively reduces the distance between similar pairs while increasing it between dissimilar ones which helps refine the model's ability to discriminate between relevant features. 
To ensure faster performance at runtime, these embeddings are calculated and stored in Chroma dB, 
an AI-native vector database designed to efficiently handle high-dimensional embeddings which allows for faster similarity search compared to traditional databases~\cite{chroma_db}.

\subsection{Retrieval with Re-ranking}

The performance of \ac{RAG} is highly dependent on the relevance and quality of the retrieved context. 
This work employs a cross-encoder re-ranker, a widely adopted semantic re-ranking method to ensure that the most relevant contexts are ranked first \cite{huggingface_msmarco_minilm_l_6_v2}.
% This work leverages a re-ranking algorithm to ensure that the most relevant contexts are ranked first.
% We use a cross-encoder reranker \cite{Reimers2019SentenceBERTSE},. 
Unlike bi-encoders, which embed each chunk independently, cross-encoders process pairs of text to calculate the similarity between them. This approach, allows it to fully capture the interactions and relationships between the query and each chunk of context. We specifically use the \textit{ms-marco-MiniLM-L-6-v2} model 
% In this paper, we utilize the "cross-encoder/ms-marco-MiniLM-L-6-v2" reranker 
due to its balance between efficiency and performance, both essential for telecom tasks \cite{huggingface_msmarco_minilm_l_6_v2}
% \footnote{https://huggingface.co/cross-encoder/ms-marco-MiniLM-L-6-v2}. 

\subsection{Extending the Context Window with SelfExtend}

% \textcolor{red}{It is also wrong to assume that all users of the telecom oracle are proficient in prompting \ac{LM}. For instance, inexperienced users might use queries that might be vague, ambiguous, or lack the specificity needed to extract the most relevant information. Providing the \ac{LM} with a wide array of context could allow it to sufficiently address the user's query. For instance, \texttt{"From a UE perspective, what does the UE need to be aware of?"} and \texttt{"How are messages routed in the 5G system?"} are an inherently broad question. It could touch upon numerous aspects such as network registration, protocol compliance, and so on. To adequately respond the generator needs to be supplemented with a substantial number of chunks.}

% can we start a section with "however"?
\Acp{SLM} typically struggle to generalize effectively to input sequences longer than those encountered during training. This presents as a challenge during inference with long contexts.
% which presents challenges during  
The context window of \acp{SLM} are often short. For example, Phi-2 has a context window of 2048 tokens. Semantic chunking does not guarantee a fixed chunk size, and therefore, might exceed the context window of the \acp{SLM}. 

% broad questions and broad user requests
% for instance, \texttt{"From a UE perspective, what does the UE need to be aware of?"} and \texttt{"How are messages routed in the 5G system?"} are an inherently broad question. It could touch upon numerous aspects such as network registration, protocol compliance, and so on. To adequately respond the generator needs to be supplemented  with a substantial number of chunks.

% It is also wrong to assume that all users of the telecom oracle are proficient in prompting \ac{LM}. For instance, inexperienced users might use queries that might be vague, ambiguous, or lack the specificity needed to extract the most relevant information. Providing the \ac{LM} with a wide array of context could allow it to sufficiently address the user's query.

Furthermore, supplementing the context with tables from telecom documents is crucial given the significance of the information they hold. Therefore, in order to support enhanced performance, and to allow for future expansions, we utilize SelfExtend\cite{self_extend}.

This method leverages the inherent capabilities of LLMs to handle extended contexts without the need for fine-tuning. SelfExtend achieves this by implementing a bi-level attention mechanism: grouped attention for capturing dependencies between distant tokens, and neighbor attention for focusing on adjacent tokens. These attentions are computed using the model's existing self-attention during inference. By making use of SelfExtend, we are able to extend Phi-2's context window to 8192 tokens.

\subsection{The Generator: Fine-tuned Phi-2 with Multiple Contexts}

A substantial portion of telecom key terms and special language are
confined to specification documents and white papers, neither of which \acp{LM} are heavily trained on. We fine-tune Phi-2 to overcome this and to adapt the model to recognize telecom terminology.

It should also be emphasized that fine-tuning the model is not intended to expand its knowledge base, but rather to enhance its ability to discern important details within the context and respond in the correct format. 
To accommodate limited resources, gradient accumulation and \ac{LoRA} are utilized for fine-tuning~\cite{Edward_2021_LoRA}. 
Gradient accumulation is a technique that allows the model to effectively handle large batch sizes by minimizing the memory needed for storing gradients. It does this by processing several small batches and accumulating the gradients from each batch before updating the weights rather than calculating and updating the weights after each batch. 

% \ac{LoRA} and its variation \ac{QLoRA} are described in the subsections below.
% 

% \subsubsection{QLoRA/LoRA}

\begin{figure}[htbp]
    \centering
    \includegraphics
    [width=0.55\columnwidth]{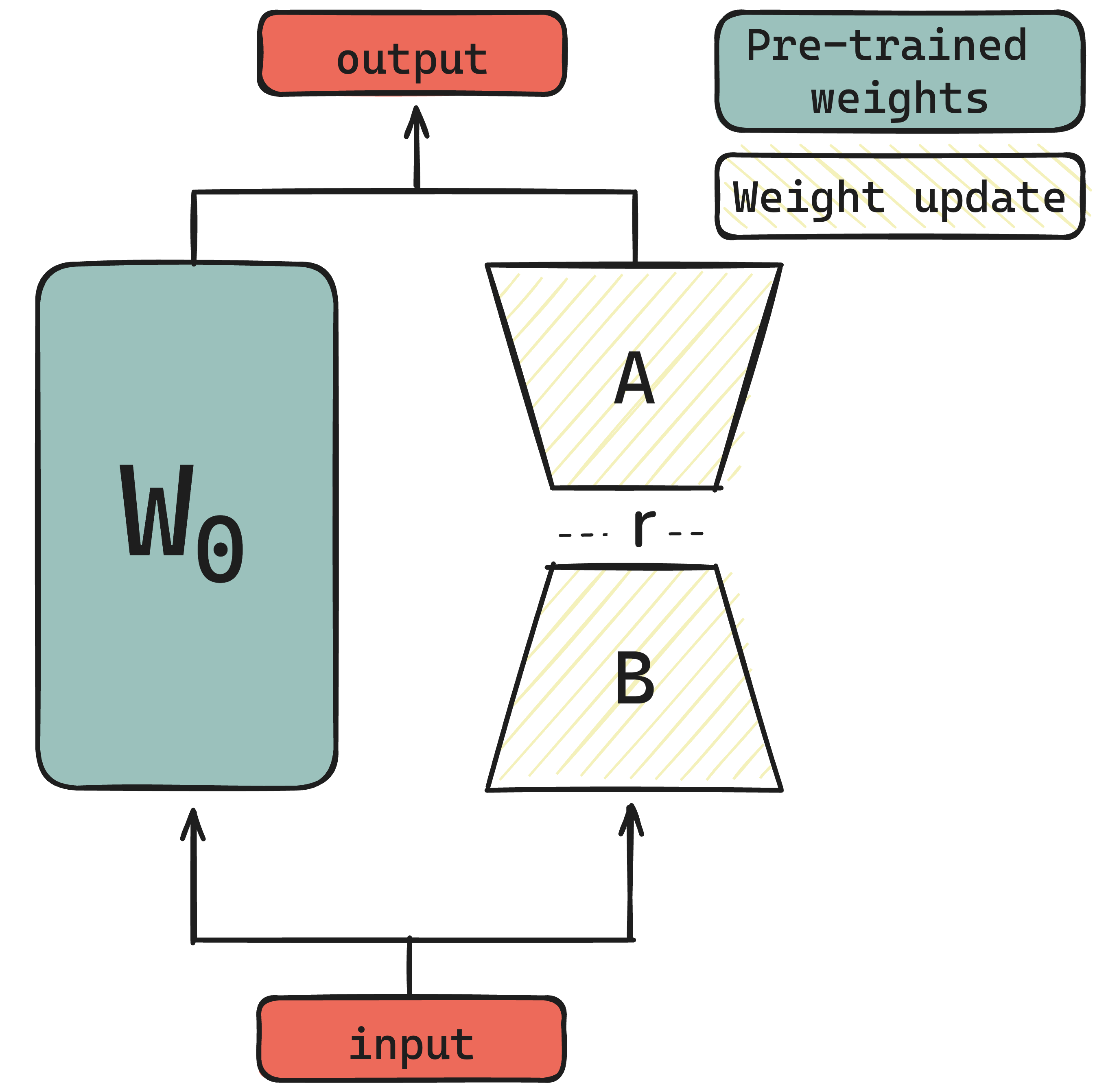}
    % \includesvg{some_svg_file}
    % \includegraphics
    % [width=0.6\linewidth]{figures/LoRA_arch_v1.png}
    \caption{Schematic illustration of the Low-Rank Adaptation (LoRA) technique for efficient fine-tuning of neural networks with low-rank matrices (e.g., \acp{LM}).}
    \label{fig:LoRA}
\end{figure}

% \newpage
% \begin{figure}[htbp]
%     \centering
%     \includegraphics
%     [width=0.65\columnwidth]{converted/LoRA_fixed.pdf}
%     % \includesvg{some_svg_file}
%     % \includegraphics
%     % [width=0.6\linewidth]{figures/LoRA_arch_v1.png}
%     \caption{Schematic illustration of the Low-Rank Adaptation (LoRA) technique for efficient fine-tuning of neural networks with low-rank matrices (e.g., \acp{LM}).}
%     \label{fig:LoRA}
% \end{figure}

% However, fine-tuning all the parameters in a model is expensive and resource-intensive deeming it unfeasible. 
Fine-tuning all parameters in a model is impractical in our domain due to resource constraints. Moreover, for smaller or specialized datasets, this approach risks overfitting and poor generalization, potentially yielding diminishing returns.
% basic idea

The concept underlying \ac{LoRA} is that pre-trained \acp{LM} possess a low `intrinsic dimension' \cite{Aghajanyan2020}. That is, the model's essential information is concentrated in a smaller subspace, even if the overall parameter space is high-dimensional. \ac{LoRA} harnesses this observation and focus the model updates on a smaller only a subset of the learning parameters as illustrated in Fig. \ref{fig:LoRA}.
% Building on this concept, \ac{LoRA} enables the model to efficiently learn, even when only a subset of the parameters are being trained. 
During fine-tuning, weight updates $W_{new}$ are represented as 
\begin{equation}
W_{new} = W_0 + \Delta W,
\label{eq:weight_update}
\end{equation}
where $W_0 \in \mathcal{R}^{d\times k}$ are the initial pre-trained weights, and $\Delta W \in \mathcal{R}^{d\times k}$ represents the change in weights. Computing $\Delta W$ is computationally expensive. In \ac{LoRA}, trainable parameters $\Delta W$ are expressed as a product of two low-rank matrices, $B\times A$, where $B\in \mathcal{R}^{d\times r}$  and $A\in \mathcal{R}^{r\times k}$, with rank $r \ll \min(d,k)$. Thus, with \ac{LoRA}, weight updates $W_{new}$ are computed as 
\begin{equation}
W_{new} = W_0 + (\frac{\alpha}{r}) B\times A,
\label{eq:weight_update}
\end{equation}
where $\alpha$ is a scaling factor, reflecting how important the weight updates are to the initial pre-trained weights.
As a result of this matrix decomposition, only $d \times r + r \times k $ parameters need to be updated.

\subsection{Prompt Engineering}
Prompt engineering is crucial for optimizing the performance of \acp{SLM}
like Phi-2 in telecom applications.
% The way Phi-2 is prompted affects the quality and relevance of its output. 
In this work, we supplement the inputted question with chunks of relevant context, and a set of instructions for the \acp{SLM} to follow, forming the prompt. An exemplary prompt is outlined in \ref{fig:prompt}. This approach helps unify the format of the \ac{SLM}'s output, focus its attention on the appearance of critical terms, and encourage it to reply on the context rather than its prior knowledge.
This prompt results in focused outputs, ensuring accurate and relevant responses for complex telecom inquiries.

\section{Experimental results}\label{section_simulations}
\subsection{Settings}

% \begin{table}[htbp]
%     \centering
%     \caption{\textbf{temp caption, Fine-tuning hyper-parameter}}
%     \label{tab:finetuning_para}
%     \includegraphics{figures/hyperparameters.pdf}
% \end{table}

% \begin{table}[htbp]
%     \centering
%     \caption{TODO: fix caption. The table uses the following acronyms: VR for Vector Retriever}
%     \label{tab:eval}
%     \includegraphics{figures/exp_para_v2.pdf}
% \end{table}

% Describe the dataset (training, test and 3GPP doc).

The RAG framework relies on around 550 \ac{3GPP} documents up to Release 18. For fine-tuning, we utilize the TeleQnA dataset which contains 10,000 multiple-choice questions~\cite{maatouk2023teleqna}. The testing set features another 2,000 multiple-choice questions that focus on \ac{3GPP} standards. For fine-tuning, we set the following parameters: weight decay of 0.01, batch size of 32, dropout rate of 0.05, and learning rate of $10^{-4}$. For \ac{LoRA}, we set the rank to 32 and the alpha to 64. The fine-tuning tagets the following layers: 'q\_proj','k\_proj','v\_proj', and 'dense'.
For semantic chunking, there are two hyper-parameters, namely the breakpoint percentile threshold and the buffer size. The former represents the percentile of cosine dissimilarity that must be exceeded between a group of sentences and the next to form a node. The latter determines the number of sentences to group together when evaluating semantic similarity. We set the breakpoint percentile threshold to 90, and the buffer size to 3.

For the vector retriever, for each query, we first retrieve 150 chunks, from which the re-ranker would return the top 15 most relevant chunks. The aforementioned hyper-parameters are not necessarily optimized, but rather set based on qualitative judgment and limited exploration.
Fine-tuning the Phi-2 generator (with multiple contexts) relies on three retrieved-context chunks along with the set of instructions and the respective query. For reproducibility and reuse, our source code is made publicly available \cite{repo}.

\subsection{Results and Analysis}
\begin{table}[htbp]
    \centering
    \caption{Accuracy comparison of ours fine-tuned Phi-2 model against baseline models, both with and without retrieved context.}
    \label{tab:models_eval}
    \includegraphics{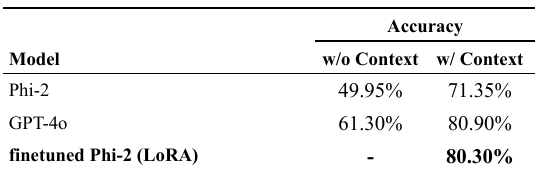}
\end{table}

\begin{table}[]
    \centering
    \caption{Performance comparison of various configurations of the fine-tuned Phi-2 model with RAG and additional components; the table uses the following acronyms: SE for SelfExtend, RR for Rerank, SC for Semantic Chunking, and MC for Multiple Context.}
    \label{tab:components}
    \includegraphics{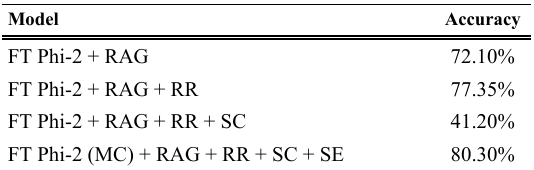}
\end{table}

% very llm, but the data is not disclosed
% check for paper that approximate ("reported to app. this bb)

In Table \ref{tab:models_eval}, we benchmark our developed framework against three other solutions: base Phi-2 (2.7B) and GPT4o, a state-of-the-art model with extensive capacity. 
% belongs to the class of large-scale language models, known for having billions of parameters,
As expected, the base Phi-2 model shows poor performance of about 49.95\% accuracy when tested on the dataset when no context is provided. Although the performance is notably improved to 71.35\% when the base model is supplemented by retrieved-context, it is still outperformed by our proposed model. Our fine-tuned RAG model is better aligned with the required task, enabling it to leverage the retrieved context better and produce more accurate recommendations.

% Since 
% GPT's performance is not consistent with its size. This could be attributed to
Notably, our developed framework performs on par with GPT-4o with and without context. Interestingly, the size of GPT-4o does not directly correlate with its performance on this specialized task. One possible explanation for this is that GPT-4o, being a highly generalized model, lacks precise alignment with the specific task requirements.
% correspond proportionally to its performance.
% One potential reason for this is that
% GPT-4o is a significantly more generalized model, therefore it is not well-aligned to the specialized task at hand. 
Qualitative analysis suggests that GPT-4o's \textit{a priori} knowledge from other domains sometimes interferes with our domain-specific task, albeit when the relevant context is present in the prompt. 

% It is also important to note that though GPT-4o mini is small, fine-tuning it and deploying the fine-tuned version is expensive, further supporting the case for using an open-source model such as Phi-2.
% \textcolor{red}{I'm not sure why gtp-4o is preforming significantly worse than gpt4o -min, might need to check what it outputs}

% \begin{table}[htbp]
%     \centering
%     \caption{The table uses the following acronyms: SE for selfExtend, RR for rerank, SC for Semantic Chunking, and MC for Multiple Context.}
%     \label{tab:components}
%     \includegraphics{figures/prompt.png}
% \end{table}

% We also note that the model fine-tuned using \ac{QLoRA} achieves comparable performance to that of the model fine-tuned using \ac{LoRA}. When loaded as a 4-bit model during inference, the performance does degrade slightly. 

% Quantizing a model can save memory and computational resources, but it can also introduce small inaccuracies in the model's representations, leading to a minor drop in performance. The fact that this degradation is only slight indicates that \ac{QLoRA} and \ac{LoRA} are robust to quantization, maintaining much of their effectiveness even in a more compressed form.

Table \ref{tab:components} represents a brief ablation study, where we analyze the impact of each added component on the predictive accuracy.
% analyzes the effect of each component added to our framework.
The re-ranking algorithm has significant positive effects, adding 5\% in accuracy. This underscores the importance of prioritizing retrieved-context chunks. When implemented on its own, semantic chunking degrades the performance. However, when paired with SelfExtend, the accuracy is significantly increased to 80.30\% which is an increase of around 8\% compared to the base model. Although semantic chunking produces more semantically coherent chunks (as shown in Fig. \ref{fig:sem_chunking}), it often results in longer chunks, and SelfExtend helps incorporate them into complete semantic units.
% Furthermore, our results demonstrate the effectiveness of SelfExtend when questions are under-specified or open-ended. 

\section{Conclusions And Future Works}\label{section_conclusions}
The significance of \acp{SLM} has not gone unrecognized in the industry, as a shift from \acp{LLM} to \acp{SLM} can be evidenced by the recent directions of several big firms. They are especially pertinent in the telecom industry given the constraints often encountered in AI-driven telecom systems, such as the need for efficient deployment on edge devices with limited computational power. Coupled with RAG, \acp{SLM} have the potential to be a dominant tool in the industry. In this paper, we develop an fine-tuned Phi-2-based RAG system to serve as an oracle for telecommunication networks. The proposed system integrates semantic chunking  
% hybrid retrieval, 
and re-ranking to improve the relevance and accuracy of retrieved contexts. Moreover, we fine-tune the generator with a carefully designed prompt and retrieved contexts to adapt it to the problem domain. Additionally, we utilize SelfExtend to significantly extend the model's context window, enabling it to process longer sequences without fine-tuning. Experimental results show that our approach competes with larger state-of-the-art LLMs and also offers significant efficiency advantages, making it suitable for deployment on edge devices. Our approach prioritizes transparency and explainability, offering a framework that allows for scrutiny, understanding, and further development in this field. Maintaining transparency through open-source models is particularly important in telecom, where it fosters trust and facilitates community contributions.

We believe the developed system can serve as a foundation for other downstream telecommunication tasks. Future work can also explore optimizing the embedding model, better integration of structured data such as tables and graphs, and further enhancements to the RAG framework to continue advancing the capabilities of \acp{LLM} in telecom applications.

\bibliographystyle{IEEEtran}
\bibliography{ref}

% Generated by IEEEtran.bst, version: 1.14 (2015/08/26)
\begin{thebibliography}{10}
\providecommand{\url}[1]{#1}
\csname url@samestyle\endcsname
\providecommand{\newblock}{\relax}
\providecommand{\bibinfo}[2]{#2}
\providecommand{\BIBentrySTDinterwordspacing}{\spaceskip=0pt\relax}
\providecommand{\BIBentryALTinterwordstretchfactor}{4}
\providecommand{\BIBentryALTinterwordspacing}{\spaceskip=\fontdimen2\font plus
\BIBentryALTinterwordstretchfactor\fontdimen3\font minus
  \fontdimen4\font\relax}
\providecommand{\BIBforeignlanguage}[2]{{%
\expandafter\ifx\csname l@#1\endcsname\relax
\typeout{** WARNING: IEEEtran.bst: No hyphenation pattern has been}%
\typeout{** loaded for the language `#1'. Using the pattern for}%
\typeout{** the default language instead.}%
\else
\language=\csname l@#1\endcsname
\fi
#2}}
\providecommand{\BIBdecl}{\relax}
\BIBdecl

\bibitem{bubeck2023sparks}
S.~Bubeck \emph{et~al.}, ``Sparks of artificial general intelligence: Early
  experiments with {GPT}-4,'' \emph{CoRR}, vol. abs/2303.12712, 2023.

\bibitem{maatouk2024large}
A.~Maatouk, N.~Piovesan, F.~Ayed, A.~D. Domenico, and M.~Debbah, ``Large
  language models for telecom: Forthcoming impact on the industry,'' \emph{IEEE
  Commun. Mag}, pp. 1--7, 2024.

\bibitem{zhang2024interactive}
R.~Zhang, H.~Du, Y.~Liu, D.~Niyato, J.~Kang, S.~Sun, X.~Shen, and H.~V. Poor,
  ``Interactive {AI} with retrieval-augmented generation for next generation
  networking,'' \emph{IEEE Netw}, pp. 1--1, 2024.

\bibitem{largelanguagemodeldrivencurriculum}
O.~Erak, O.~Alhussein, S.~Naser, N.~Alabbasi, D.~Mi, and S.~Muhaidat, ``Large
  language model-driven curriculum design for mobile networks,'' \emph{CoRR},
  vol. abs/2405.18039, 2024.

\bibitem{yao2024largelanguagemodelunlearning}
Y.~Yao, X.~Xu, and Y.~Liu, ``Large language model unlearning,'' \emph{CoRR},
  vol. abs/2310.10683, 2024.

\bibitem{huang2023_survey_hallucination_LLMs}
L.~Huang, W.~Yu, W.~Ma, W.~Zhong, Z.~Feng, H.~Wang, Q.~Chen, W.~Peng, X.~Feng,
  B.~Qin, and T.~Liu, ``A survey on hallucination in large language models:
  Principles, taxonomy, challenges, and open questions,'' \emph{CoRR}, vol.
  abs/2311.05232, 2023.

\bibitem{piovesan2024_Telecom_LLMs_large}
N.~Piovesan, A.~D. Domenico, and F.~Ayed, ``Telecom language models: Must they
  be large?'' \emph{CoRR}, vol. abs/2403.04666, 2024.

\bibitem{PHI2}
M.~Javaheripi and S.~Bubeck, ``\BIBforeignlanguage{en}{Phi-2: The surprising
  power of small language models},''
  \url{https://www.microsoft.com/en-us/research/blog/phi-2-the-surprising-power-of-small-language-models/},
  Dec. 2023, accessed: 2024-8-19.

\bibitem{gemini}
G.~Team \emph{et~al.}, ``Gemini: A family of highly capable multimodal
  models,'' \emph{CoRR}, vol. abs/2312.11805, 2024.

\bibitem{llama}
H.~Touvron \emph{et~al.}, ``Llama 2: Open foundation and fine-tuned chat
  models,'' \emph{CoRR}, vol. abs/2307.09288, 2023.

\bibitem{maatouk2023teleqna}
A.~Maatouk, F.~Ayed, N.~Piovesan, A.~D. Domenico, M.~Debbah, and Z.-Q. Luo,
  ``Tele{Q}n{A}: A benchmark dataset to assess large language models
  telecommunications knowledge,'' \emph{CoRR}, vol. abs/2310.15051, 2023.

\bibitem{bornea2024_telco_RAG}
A.-L. Bornea, F.~Ayed, A.~D. Domenico, N.~Piovesan, and A.~Maatouk,
  ``Telco-{RAG}: Navigating the challenges of retrieval-augmented language
  models for telecommunications,'' \emph{CoRR}, vol. abs/2404.15939, 2024.

\bibitem{self_extend}
H.~Jin, X.~Han, J.~Yang, Z.~Jiang, Z.~Liu, C.-Y. Chang, H.~Chen, and X.~Hu,
  ``{LLM Maybe LongLM: Self-Extend LLM Context Window Without Tuning},''
  \emph{CoRR}, vol. abs/2401.01325, 2024.

\bibitem{karim2023_SPEC5G}
I.~Karim, K.~S. Mubasshir, M.~M. Rahman, and E.~Bertino, ``{SPEC5G}: A dataset
  for {5G} cellular network protocol analysis,'' in \emph{Proc. IJCNLP-AACL},
  2023, pp. 20--38.

\bibitem{karapantelakis2024_TeleRoBERTa}
A.~Karapantelakis, M.~Thakur, A.~Nikou, F.~Moradi, C.~Orlog, F.~Gaim, H.~Holm,
  D.~D. Nimara, and V.~Huang, ``Using large language models to understand
  telecom standards,'' \emph{CoRR}, vol. abs/2404.02929, 2024.

\bibitem{nikbakht2024_TSPEC}
R.~Nikbakht, M.~Benzaghta, and G.~Geraci, ``{TSpec-LLM}: An open-source dataset
  for {LLM} understanding of {3GPP} specifications,'' \emph{CoRR}, vol.
  abs/2406.01768, 2024.

\bibitem{gajjar2024_ORAN_Bench_13K}
P.~Gajjar and V.~K. Shah, ``{ORAN-Bench-13K}: An open source benchmark for
  assessing {LLM}s in open radio access networks,'' \emph{CoRR}, vol.
  abs/2407.06245, 2024.

\bibitem{ahmed2024linguisticintelligencelargelanguage}
T.~Ahmed, N.~Piovesan, A.~D. Domenico, and S.~Choudhury, ``Linguistic
  intelligence in large language models for telecommunications,'' \emph{CoRR},
  vol. abs/2402.15818, 2024.

\bibitem{bariah2023understandingtelecomlanguagelarge}
L.~Bariah, H.~Zou, Q.~Zhao, B.~Mouhouche, F.~Bader, and M.~Debbah,
  ``Understanding telecom language through large language models,'' in
  \emph{Proc. IEEE Globecom}, 2023, pp. 6542--6547.

\bibitem{Liu_LlamaIndex_2022}
\BIBentryALTinterwordspacing
J.~Liu, ``{LlamaIndex},'' 2022. [Online]. Available:
  \url{https://github.com/jerryjliu/llama_index}
\BIBentrySTDinterwordspacing

\bibitem{bge_embedding}
S.~Xiao, Z.~Liu, P.~Zhang, N.~Muennighoff, D.~Lian, and J.-Y. Nie, ``{C-Pack}:
  Packaged resources to advance general chinese embedding,'' \emph{CoRR}, vol.
  abs/2309.07597, 2024.

\bibitem{chen2020_contrastive}
T.~Chen, S.~Kornblith, M.~Norouzi, and G.~Hinton, ``A simple framework for
  contrastive learning of visual representations,'' in \emph{Proc. ICML}, 2020,
  pp. 1597--1607.

\bibitem{chroma_db}
{Chroma}, ``{ChromaDB},'' \url{https://www.trychroma.com/}, accessed:
  08/05/2024.

\bibitem{huggingface_msmarco_minilm_l_6_v2}
{Inferless}, ``Ms marco: ms-marco-minilm-l-6-v2,''
  \url{https://huggingface.co/cross-encoder/ms-marco-MiniLM-L-6-v2}, 2023,
  accessed: 2024-08-18.

\bibitem{Edward_2021_LoRA}
E.~J. Hu, yelong shen, P.~Wallis, Z.~Allen-Zhu, Y.~Li, S.~Wang, L.~Wang, and
  W.~Chen, ``Lo{RA}: Low-rank adaptation of large language models,'' in
  \emph{Proc. ICLR}, 2022.

\bibitem{Aghajanyan2020}
A.~Aghajanyan, S.~Gupta, and L.~Zettlemoyer, ``Intrinsic dimensionality
  explains the effectiveness of language model fine-tuning,'' in \emph{Proc.
  ACL-IJCNLP}, 2021, pp. 7319--7328.

\bibitem{repo}
N.~Alabbasi and O.~Erak, ``Specializing large language models for telecom
  networks,''
  \url{https://github.com/Nouf-Alabbasi/oKUmura_AI_Telecom_challenge}, 2024.

\end{thebibliography}

\end{document}